\documentclass[10pt,twocolumn,letterpaper]{article}
\usepackage{cvpr}
\usepackage{times}
\usepackage{graphicx}
\usepackage{amsmath}
\usepackage{amssymb}
\usepackage{lipsum}
\usepackage[pagebackref=true,breaklinks=true,%
letterpaper=true,colorlinks,bookmarks=false]{hyperref}
\usepackage{cleveref}
\usepackage{siunitx}

\newcommand{\bx}{\mathbf{x}}

\newcommand{\bw}{\mathbf{w}}

\newcommand{\fns}[1]{\footnotesize{#1}}
\newcommand{\IoUset}{\operatorname{IoU}_\text{set}}
\newcommand{\IoUind}{\operatorname{IoU}_\text{ind}}
\newcommand\blfootnote[1]{%
  \begingroup
  \renewcommand\thefootnote{}\footnote{#1}%
  \addtocounter{footnote}{-1}%
  \endgroup
}
\graphicspath{{./figures/}}

\cvprfinalcopy
\def\cvprPaperID{4217}

\ifcvprfinal\fi
\title{Net2Vec: Quantifying and Explaining how Concepts are Encoded by Filters in Deep Neural Networks}
\author{Ruth Fong\\
University of Oxford\\
{\tt\small ruthfong@robots.ox.ac.uk}
\and
Andrea Vedaldi\\
University of Oxford\\
{\tt\small vedaldi@robots.ox.ac.uk}}

\begin{document}
\maketitle
\begin{abstract}
In an effort to understand the meaning of the intermediate representations captured by deep networks, recent papers have tried to associate specific semantic concepts to individual neural network filter responses, where interesting correlations are often found, largely by focusing on extremal filter responses. In this paper, we show that this approach can favor easy-to-interpret cases that are not necessarily representative of the average behavior of a representation.

A more realistic but harder-to-study hypothesis is that semantic representations are distributed, and thus filters must be studied in conjunction. In order to investigate this idea while enabling systematic visualization and quantification of multiple filter responses, we introduce the Net2Vec framework, in which semantic concepts are mapped to vectorial embeddings based on corresponding filter responses. By studying such embeddings, we are able to show that 1., in most cases, multiple filters are required to code for a concept, that 2., often filters are not concept specific and help encode multiple concepts, and that 3., compared to single filter activations, filter embeddings are able to better characterize the meaning of a representation and its relationship to other concepts.
\end{abstract}

\setlength{\tabcolsep}{2pt}
\section{Introduction}\label{s:intro}

While deep neural networks keep setting new records in almost all problems in computer vision, our understanding of these black-box models remains very limited. Without developing such an understanding, it is difficult to characterize and work around the limitations of deep networks, and improvements may only come from intuition and trial-and-error.

\begin{figure}
  \centering
  \includegraphics[width=\linewidth]{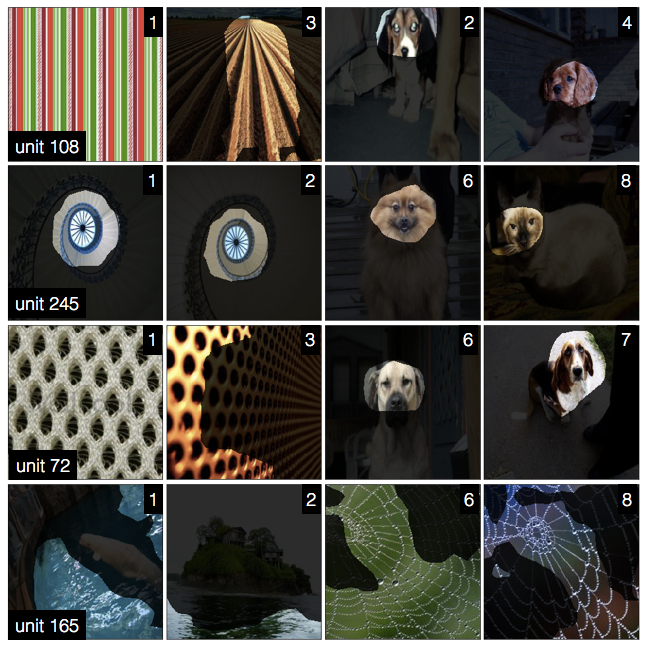}
  \caption{The diversity of BRODEN~\cite{bau2017network} images that most activate certain AlexNet conv5 filters motivates us to investigate to what extent a single filter encodes a concept fully, without needing other units, and exclusively, without encoding other concepts. An image's corner number $n$ denotes that it is the $n$-th most maximally activating image for the given filter. Masks were generated by our slightly modified NetDissect~\cite{bau2017network} approach (\cref{s:seg-single}) and are upsampled first before thresholding for smoothness.}
  \label{f:splash}
\end{figure}

For deep learning to mature, a much better theoretical and empirical understanding of deep networks is thus required. There are several questions that need answering, such as how a deep network is able to solve a problem such as classifying an image, or how it can generalize so well despite having access to limited training data in relation to its own capacity~\cite{zhang16understanding}. In this paper, we ask in particular \emph{what a convolutional neural network has learned to do} once training is complete. A neural network can be seen as a sequence of functions, each mapping an input image to some intermediate representation. While the final output of a network is usually easy to interpret (as it provides, hopefully, a solution to the task that the network was trained to solve), the meaning of the intermediate layers is far less clear. Understanding the information carried by these representations is a first step to understanding how these networks work.

Several authors have researched the possibility that individual filters in a deep network are responsible for capturing particular semantic concepts. The idea is that low-level primitives such as edges and textures are recognized by earlier layers, and more complex objects and scenes by deeper ones. An excellent representative of this line of research is the recent Network Dissection approach by~\cite{bau2017network}. The authors of this paper introduce a new dataset, BRODEN, which contains pixel-level segmentation for hundreds of low- and high-level visual concepts, from textures to parts and objects. They then study the correlation between extremal filter responses and such concepts, seeking for filters that are strongly responsive for particular ones.

While this and similar studies~\cite{zhou2014object,Zeiler13,le2013building} did find clear correlations between feature responses and various concepts, such an interpretation has intrinsic limitations. This can be seen from a simple counting argument: the number of available feature channels is usually far smaller than the number of different concepts that a neural network may need to encode to interpret a complex visual scene. This suggests that, at the very least, the representation must use combinations of filter responses to represent concepts or, in other words, be at least in part distributed.

\paragraph{Overview.} The goal of this paper is to go beyond looking at individual filters, and to study instead what information is captured by \textbf{combinations} of neural network filters. In this paper, we conduct a thorough analysis to investigate how semantic concepts, such as objects and their parts, are encoded by CNN filters. In order to make this analysis manageable, we introduce the Net2Vec framework (\cref{s:net2vec}), which aligns semantic concepts with filter activations. It does so via learned concept embeddings that are used to weight filter activations to perform semantic tasks like segmentation and classification. Our concept vectors can be used to investigate both quantitatively \emph{and} qualitatively  the ``overlap'' of filters and concepts. Our novelty lies in outlining methods that go beyond simply demonstrating that multiple filters better encode concepts that single ones~\cite{agrawal2014analyzing,wang2015unsupervised} to quantifying and describing how a concept is encoded. Principally, we gain unique, interpretive power by formulating concepts vectors as embeddings.

Using Net2Vec, we look first at two questions (\cref{s:cover}):
(1) To what extent are individual filters sufficient to express a concept? Or, are multiple filters required to code for a single concept?
(2) To what extent does a filter exclusively code for a single concept? Or, is a filter shared by many, diverse concepts?
While answers to these questions depend on the specific filter or concept under consideration, we demonstrate how to \textbf{quantify} the ``overlap'' between filters and concepts and show that there are many cases in which both notions of exclusive overlap do not hold. That is, if we were to interpret semantic concepts and filter activations as corresponding set of images, in the resulting Venn's diagram the sets would intersect partially but neither kind of set would contain or be contained by the other.

While quantifying the relationship between concepts and representation may seem an obvious aim, so far much of the research on explaining how concepts are encoded by deep networks roughly falls into two more qualitative categories:
(1) Interpretable visualizations of how single filters encode semantic concepts;
(2) Demonstrations of distributive encoding with limited explanatory power of how a concept is encoded.
In this work, we present methods that seek to marry the interpretive benefits of single filter visualizations with quantitative demonstrations of how concepts are encoded across multiple filters (\cref{s:interp}). 

As part of our analysis, we also highlight the problem with visualizing only the inputs that maximally activate a filter and propose evaluating the power of explanatory visualizations by how well they can explain the whole distribution of filter activations (\cref{s:nonmax}).

\section{Related Work}\label{s:related}

\paragraph{Visualizations.} Several methods have been proposed to explain what a single filter encodes by visualizing a real~\cite{Zeiler13} or generated~\cite{le2013building,Simonyan14a,nguyen2016synthesizing} input that most activates a filter; these techniques are often used to argue that single filters substantially encode a concept. In contrast,~\cite{szegedy2013intriguing} shows that visualizing the real image patches that most activate a layer's filters after a random basis has been applied also yields semantically, coherent patches.~\cite{zhou2014object,bau2017network} visualize segmentation masks extracted from filter activations for the most confident or maximally activating images; they also evaluate their visualizations using human judgments.

\paragraph{Distributed Encodings.}\cite{agrawal2014analyzing} demonstrates that most PASCAL classes require more than a few hidden units to perform classification well. Most similar to~\cite{zhou2014object,bau2017network},~\cite{gonzalez2016semantic} concludes that only a few hidden units encode semantic concepts robustly by measuring the overlap between image patches that most activate a hidden unit with ground truth bounding boxes and collecting human judgments on whether such patches encode systematic concepts.~\cite{wang2015unsupervised} compares using individual filter activations with using clusters of activations from all units in a layer and shows that their clusters yielded better parts detectors and qualitatively correlated well with semantic concepts.~\cite{alain2016understanding} probes mid-layer filters by training linear classifiers on their activations and analyzing them at different layers and points of training.

 
\section{Net2Vec}\label{s:net2vec}

With our Net2Vec paradigm, we propose aligning concepts to filters in a CNN by (a) recording filter activations of a pre-trained network when probed by inputs from a reference, ``probe'' dataset and (b) learning how to weight the collected probe activations to perform various semantic tasks. In this way, for every concept in the probe dataset, a concept weight is learned for the task of recognizing that concept. The resulting weights can then be interpreted as concept embeddings and analyzed to understand how concepts are encoded. For example, the performance on semantic tasks when using learned concept weights that span all filters in a layer can be compared to when using only a single filter or subset of filters.

In the remainder of the section, we provide details for how we learn concept embeddings by learning to segment (\ref{s:seg}) and classify (\ref{s:class}) concepts. We also outline how we compare embeddings arising from using only a restricted set of filters, including single filters. Before we do so, we briefly discuss the dataset used to learn concepts.

\paragraph{Data.} We build on the BRODEN dataset recently introduced by~\cite{bau2017network} and use it to primarily probe AlexNet~\cite{Krizhevsky12} trained on the ImageNet dataset~\cite{russakovsky2015imagenet} as a representative model for image classification. BRODEN contains over 60,000 images with pixel- and image-level annotations for 1197 concepts across 6 categories: scenes (468), objects (584), parts (234), materials (32), textures (47), and colors (11). We exclude 8 scene concepts for which there were no validation examples. Thus, of the 1189 concepts we consider, all had image-level annotations, but only 682 had segmentation annotations, as only image-level annotations are provided for scene and texture concepts. Note that our paradigm can be generalized to any probe dataset that contains pixel- or image-level annotations for concepts. To compare the effects of different architectures and supervision, we also probe VGG16~\cite{Simonyan15} conv5\_3 and GoogLeNet~\cite{Szegedy15} inception5b trained on ImageNet~\cite{russakovsky2015imagenet} and Places365~\cite{zhou2016places} as well as conv5 of the following self-supervised, AlexNet networks: tracking~\cite{wang2015unsupervised}, audio~\cite{owens2016ambient}, objectcentric~\cite{gao2016object}, moving~\cite{agrawal2015learning}, and egomotion~\cite{jayaraman2015learning}. Post-ReLU activations are used. 


\subsection{Concept Segmentation}\label{s:seg}


In this section, we show how learning to segment concepts can be used to induce concept embeddings using either all the filters available in a CNN layer or just a single filter. We also show how embeddings can be used to quantify the degree of overlap between filter combinations and concepts. This task is performed on all $682$ Broden concepts with segmentation annotations, which excludes scene and texture concepts.

\subsubsection{Concept Segmentation by a Single Filter}\label{s:seg-single}

We start by considering single filter segmentation following~\cite{bau2017network}'s paradigm with three minor modifications, listed below. For every filter $k$, let $a_k$ be its corresponding activation (at a given pixel location and for a given input image). The $\tau = 0.005$ activation's quantile $T_k$ is determined such that $P(a_k > T_k) = \tau$, and is computed with respect to the distribution $p(a_k)$ of filter activations over all probe images and spatial locations; we use this cut-off point to match~\cite{bau2017network}.

Filter $k$ in layer $l$ is used to generate a segmentation of an image by first thresholding $A_k(\bx) > T_k$, where $A_k(\bx) \in \mathbb{R}^{H_l\times W_l}$ is the activation map of filter $k$ on input $\bx \in \mathbb{R}^{H\times W\times 3}$ and upsampling the result as needed to match the resolution of the ground truth segmentation mask $L_c(\bx)$, i.e.\ $M_k(\bx) = S(A_k(\bx) > T_k)$, where $S$ denotes a bilinear upsampling function.


Images may contain any number of different concepts, indexed by $c$. We use the symbol $\bx \in X_c$ to denote the probe images that contain concept $c$. To determine which filter $k$ best segments concept $c$, we compute a set IoU score. This score is given by the formula
\begin{equation}
\IoUset(c;M_k,s) = \frac
{\sum_{\bx \in X_{s,c}} |M_k(\bx) \cap L_c(\bx)|}
{\sum_{\bx \in X_{s,c}} |M_k(\bx) \cup L_c(\bx)|}
\label{e:iou_set}
\end{equation}
which computes the intersection over union (Jakkard index) difference between the binary segmentation masks $M_k$ produced by the filter and the ground-truth segmentation masks $L_c$. Note that sets are merged for all images in the subset $X_{s,c}$ of the data, where $s \in \{\text{train},\text{val}\}$.  The best filter $k^*(c)=\operatornamewithlimits{argmax}_k \IoUset(c;M_k,\text{train})$ is then selected on the training set and the validation score IoU $\IoUset(c;M_{k^*},\text{val})$ is reported.

We differ from~\cite{bau2017network} in the following ways: (1) we threshold before upsampling, in order to more evenly compare to the method described below; (2) we bilinearly upsample without anchoring interpolants at the center of filter receptive fields to speed up the upsampling part of the experimental pipeline; and (3) we determine the best filter for a concept on the training split $X_{\text{train},c}$ rather than $X_c$ whereas~\cite{bau2017network} does not distinguish a training and validation set.


\subsubsection{Concept Segmentation by Filter Combinations}\label{s:seg-comb}

In order to compare single-feature concept embeddings to representations that use filter combinations, we also learn to solve the segmentation task using \emph{combinations} of filters extracted by the neural network. For this, we learn weights $\bw \in \mathbb{R}^K$, where $K$ is the number of filters in a layer, to linearly combine thresholded activations. Then, the linear combination is passed through the sigmoid function $\sigma(z) = {1}/(1+\exp(-z))$ to predict a segmentation mask $M(\bx;\bw)$:
\begin{equation}\label{e:seg_linear_comb}
M(\bx;\bw) = \sigma
\left(
\sum_k w_k \cdot \mathbb{I}(A_k(\bx) > T_k)
\right)
\end{equation}
where $\mathbb{I}(\cdot)$ is the indicator function of an event. The sigmoid is irrelevant for evaluation, for which we threshold the mask predicted by $M(\bx;\bw)$ by $\frac{1}{2}$, but has an effect in training the weights $\bw$.

 

Similar to the single filter case, for each concept the weights $\bw$ are learned on $X_{\text{train},c}$ and the set IoU score computed on thresholded masks for $X_{\text{val},c}$ is reported. In addition to evaluating on the set IoU score, per-image IoU scores are computed as well:
\begin{equation}\label{e:iou_ind}
\IoUind(\bx,c;M)
= \frac{|M(\bx) \cap L_c(\bx)|}{|M(\bx) \cup L_c(\bx)|}
\end{equation}
Note that choosing a single filter is analogous to setting $\bw$ to a one-hot vector, where $w_k = 1$ for the selected filter and $w_k = 0$ otherwise, recovering the single-filter segmenter of~\cref{s:seg-single}, with the output rescaled by the sigmoid function~\eqref{e:seg_linear_comb}.

\paragraph{Training}
For each concept $c$, the segmentation concept weights $\bw$ are learned using SGD with momentum (lr $= 10^{-4}$, momentum $\gamma = 0.9$, batch size $64$, $30$ epochs) to minimize a per-pixel binary cross entropy loss weighted by the mean concept size, i.e. 1-$\alpha$:
\begin{multline}\label{e:bce_loss}
\mathcal{L}_1 = -\frac{1}{N_{s,c}} 
\sum_{\bx \in X_{s,c}}
\alpha M(\bx;\bw) L_c(\bx)
\\
+ (1 - \alpha) (1-M(\bx;\bw)(1-L_c(\bx)),
\end{multline}
where $N_{s,c} = |X_{s,c}|$, $s \in \{\text{train}, \text{val}\}$, and $\alpha = 1 - \sum_{\bx \in X_{\text{train}}} {|L_c(\bx)|}/{S}$, where $|L_c(\bx)|$ is the number of foreground pixels for concept $c$ in the ground truth (g.t.) mask for $\bx$ and $S = h_s \cdot w_s$ is the number of pixels in g.t. masks. 

\subsection{Concept Classification}\label{s:class}

As an alternate task to concept segmentation, the problem of classifying concept (i.e., to tell whether the concept occurs somewhere in the image) can be used to induce concept embeddings. In this case, we discuss first learning embeddings using generic filter combinations (\ref{s:class-comb}) and then reducing those to only use a small subset of filters (\ref{s:class-subset}).

\subsubsection{Concept Classification by Filter Combinations}\label{s:class-comb}

Similar to our segmentation paradigm, for each concept $c$, a weight vector $\bw \in \mathbb{R}^K$ and a bias term $b \in \mathbb{R}$ are learned to combine the spatially-averaged filter activations $k$; the linear combination is then passed through the sigmoid function $\sigma$ to obtain the concept posterior probability:
\begin{equation}\label{e:class_linear_comb}
f(\bx;\bw,b) = 
\sigma
\left(
b + \sum_k w_k \cdot \frac{\sum_{i=1}^{H_l}\sum_{j=1}^{W_l} A_{ijk}(\bx)}{H_l W_l}
\right)
\end{equation}
where $H_l$ and $W_l$ denote the height and width respectively of layer $l$'s activation map $A_k(\bx)$.

For each concept $c$, the training images $X_{\text{train}}$ are divided into the positive subset $X_{\text{train},c+}$ of images that contain concept $c$ and its complement $X_{\text{train},c-}$ of images that do not. While in general the positive and negative sets are unbalanced, during training, images from the two sets are sampled with equal probability in order to re-balance the data (supp. sec. 1.2). To evaluate performance, we calculate the classification accuracy over a balanced validation set.


\subsubsection{Concept Classification by a Subset of Filters}\label{s:class-subset}

In order to compare using all filters in a layer to just a subset of filters, or even individual filters, we must learn corresponding concept classifiers. Following~\cite{agrawal2014analyzing}, for each concept $c$, after learning weights $\bw$ as explained before, we choose the top $F$ by their absolute weight $|w_k|$. Then, we learn new weights $\bw' \in \mathbb{R}^F$ and bias $b'$ that are used to weight activations from only these $F$ filters. With respect to~\cref{e:class_linear_comb}, this is analogous to learning new weights $\bw' \in \mathbb{R}^K$, where $w_k' = 0$ for all filters $k$ that are not the top $F$ ones. We train such classifiers for $F \in \{1,2,3,5,10,15,20,25,30,35,40,45,50,80,100,128\}$ for the last three AlexNet layers and for all its layers for the special case $F=1$, corresponding to a single filter. For comparison, we use this same method to select subsets of filters for the segmentation task on the last layer using $F \in \{1,2,4,8,16,32,64,128,160,192,224\}$.

\section{Quantifying the Filter-Concept Overlap}\label{s:cover}

\subsection{Are Filters Sufficient Statistics for Concepts?}\label{s:filter_cover}

We start by investigating a popular hypothesis: whether concepts are well represented by the activation of individual filters or not. In order to quantify this, we consider how our learned weights, which combine information from all filter activations in a layer, compare to a single filter when being used to perform segmentation and classification on BRODEN. 

\begin{figure}[t]
\begin{center}
\includegraphics[width=\linewidth]{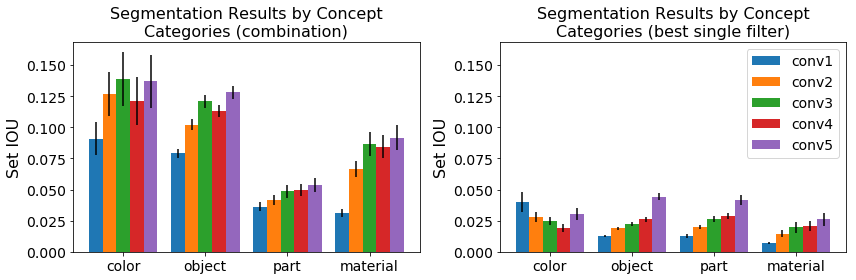}
\includegraphics[width=\linewidth]{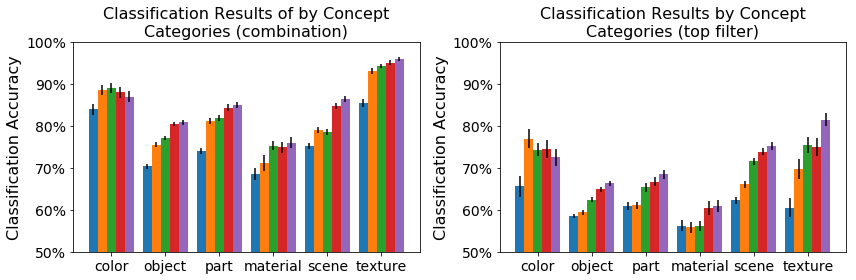}
\end{center}
\caption{Results by concept category on the segmentation (top) and classification (bottom) tasks show that, on average, using learned weights to combine filters (left) out performs using a single filter (right). Standard error is shown.}
\label{f:results_graphs}
\end{figure}

\Cref{f:results_graphs} shows that, on average, using learned weights to combine filters outperforms using a single filter on both the segmentation and classification tasks (\cref{s:seg-single,s:class-subset}) when being evaluated on validation data. The improvements can be quite dramatic for some concepts and starts in conv1. For instance, even for simple concepts like colors, filter combinations outperform individual filters by up to $4\times$ (see supp. figs. 2-4 for graphs on the performance of individual concepts). This suggests that, even if filters specific to a concept can be found, these do not optimally encode or fully ``overlap'' with the concept. In line with the accepted notion that deep layers improve representational quality, task performance generally improves as the layer depth increases, with trends for the color concepts being the notable exception. Furthermore, the average performance varies significantly by concept category and consistently in both the single- and multi-filter classification plots (bottom). This suggests that certain concepts are less well-aligned via linear combination to the filter space.

\begin{figure}[t]
\begin{center}
\includegraphics[width=0.49\linewidth]{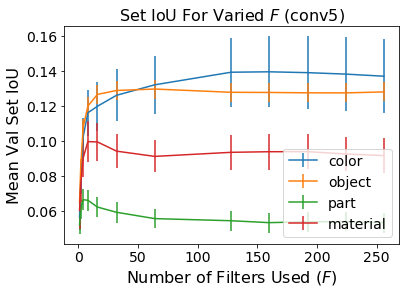}
\includegraphics[width=0.49\linewidth]{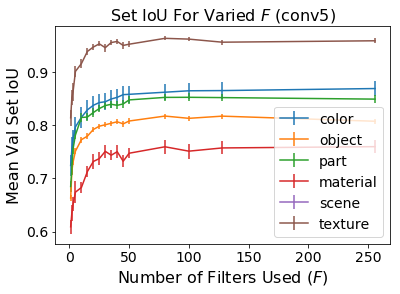}
\end{center}
\caption{Results by concept category and number of top conv5 filters used for segmentation and classification show that different categories and tasks saturate in performance at different $F$.}
\label{f:cum_num_filters}
\end{figure}

\paragraph{How many filters are required to encode a concept?} To answer this question, we observe how varying the number of top conv5 filters, $F$, from which we learn concept weights affects performance (\cref{s:class-subset}).~\Cref{f:cum_num_filters} shows that mean performance saturates at different $F$ for the various concept categories and tasks. For the classification task (right), most concept categories saturate by $F = 50$; however, scenes reaches near optimal performance around $F = 15$, which is much more quickly than that of materials. For the segmentation task (left), performance peaks much earlier at $F = 8$ for materials and parts, $F = 16$ for objects, and $F = 128$ for colors. We also observe performance drops after reaching optimal peaks for materials and parts in the segmentation class. This highlights that the segmentation task is challenging for those concept categories in particular (i.e., object parts are much smaller and harder to segment, materials are most different from network's original ImageNet training examples of objects); with more filters to optimize for, learning is more unstable and more likely to reach a sub-optimal solution.

\paragraph{Failure Cases.} While on average our multi-filter approach significantly outperforms a single-filter approach on both segmentation and classification tasks (\cref{f:results_graphs}),~\Cref{t:percent_combo_better} shows that for around $10\%$ of concepts, this does not hold. For segmentation, this percentage increases with layer depth. Upon investigation, we discovered that the concepts for which our learned weights do not outperform the best filter either have very few examples for that concept, i.e.\ mostly $|X_{\text{train},c}| \in [10,100]$ which leads to overfitting; or are very small objects, of average size less than $1\%$ of an image, and thus training with the size weighted \eqref{e:bce_loss} loss is unstable and difficult, particularly at later layers where there is low spatial resolution. A similar analysis on the classification results shows that small concept dataset size is also causing overfitting in failure cases: Of the 133 conv5 failure cases, 103 had at most 20 positive training examples and all but one had less than 100 positive training examples (supplementary material figs.\ 7 and 8).

\begin{table}[t]
\centering
\caption{Percent of concepts for which the evaluation metric (set IoU for segmentation and accuracy for classification) is equal to or better when using learned weights than the best single filter.}
\label{t:percent_combo_better}
\begin{tabular}{|c|c|c|c|c|c|}
\hline
&\fns{\textbf{conv1}}&\fns{\textbf{conv2}}&\fns{\textbf{conv3}}&\fns{\textbf{conv4}}&\fns{\textbf{conv5}}\\ \hline \hline
\fns{\textit{Segmentation}}&\fns{91.6\%}&\fns{86.8\%}&\fns{84.0\%}&\fns{82.3\%}&\fns{75.7\%}\\ \hline
\fns{\textit{Classification}}&\fns{87.8\%}&\fns{90.2\%}&\fns{85.0\%}&\fns{87.9\%}&\fns{88.1\%}\\ \hline
\end{tabular}
\end{table}

\subsection{Are Filters Shared between Concepts?}\label{s:concept_cover}
\begin{figure}
\centering
\includegraphics[width=\linewidth]{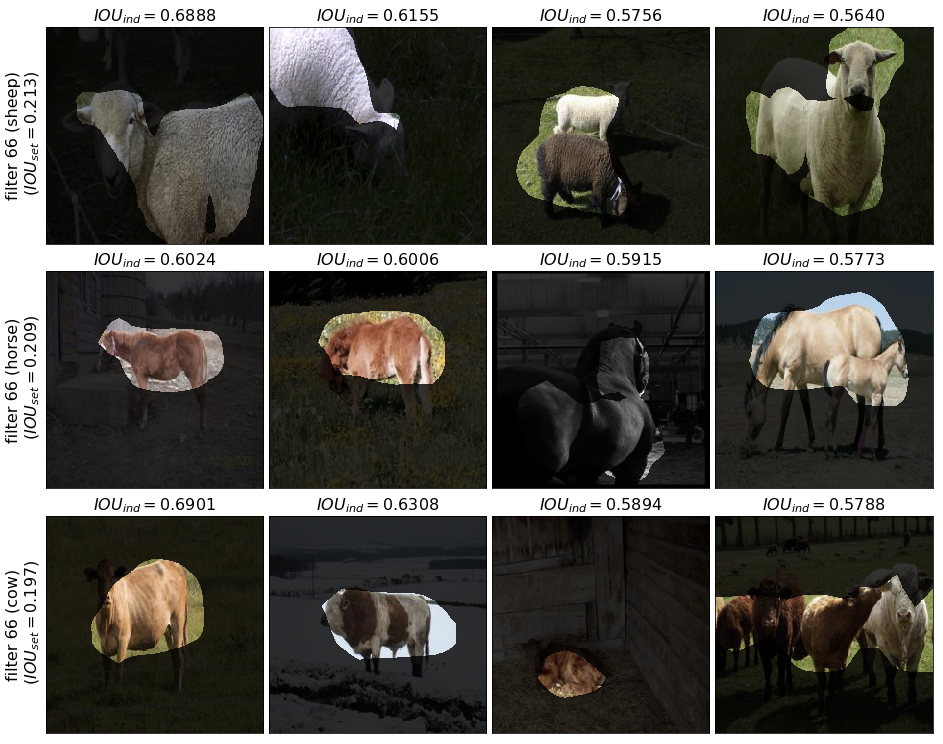}
\caption{AlexNet conv5 filter 66 appears selective for pastoral animal's torso. Validation examples for `sheep', `horse', and `cow' with the highest individual IOU scores are given (masks are upsampled before thresholding for visual smoothness).}
\label{f:mult_concept}
\end{figure}

Next, we investigate the extent to which a single filter is used to encode many concepts. Note that~\Cref{f:splash} suggests that a single filter might be activated by different concepts; often, the different concepts a filter appears to be activated by are related by a latent concept that may or may not be human-interpretable, i.e., an `animal torso' filter which also is involved in characterizing animals like `sheep', `cow', and `horse' (\cref{f:mult_concept}, supp. fig. 9).

\begin{figure}[t]
\begin{center}
\includegraphics[width=0.49\linewidth]{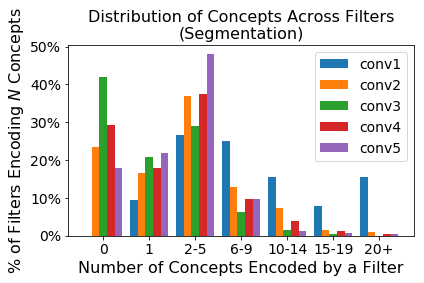}
\includegraphics[width=0.49\linewidth]{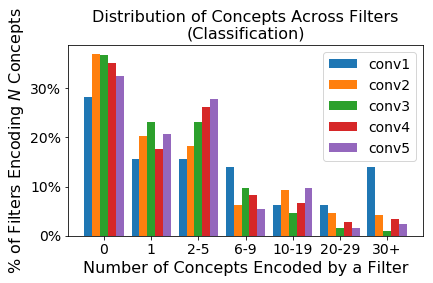}
\end{center}
\caption{For each filter in a layer, the number of concepts for which it is selected as the best filter in the segmentation (left) and classification (right) tasks is counted and binned.}
\label{f:dist_concepts_across_filters}
\end{figure}

Using the single best filters identified in both the segmentation and classification tasks, we explore how often a filter is selected as the best filter to encode a concept.~\Cref{f:dist_concepts_across_filters} shows the distribution of how many filters (y-axis) encode how many concepts (x-axis). Interestingly, around $15\%$ of conv1 filters (as well as several in all the other layers) were selected for encoding at least 20 and 30 concepts (\# of concepts / \# of conv1 filters = 10.7 and 18.6; supp. tbl. 1) for the segmentation and classification tasks respectively and a substantial portion of filters in each layer (except conv1 for the segmentation task) are never selected. The filters selected to encode numerous concepts are not exclusively ``overlapped'' by a single concept. The filters that were not selected to encode any concepts are likely not be involved in detecting highly discriminative features.

\subsection{More Architectures, Datasets, and Tasks}\label{s:more_experiments}
\begin{figure}
  \centering
  \includegraphics[width=\linewidth]{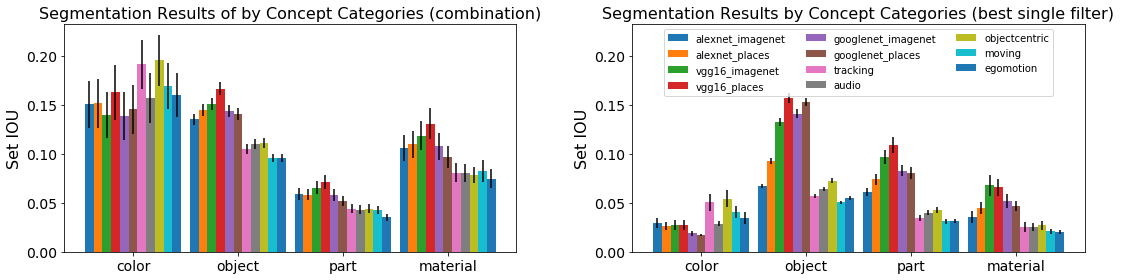}
  \includegraphics[width=\linewidth]{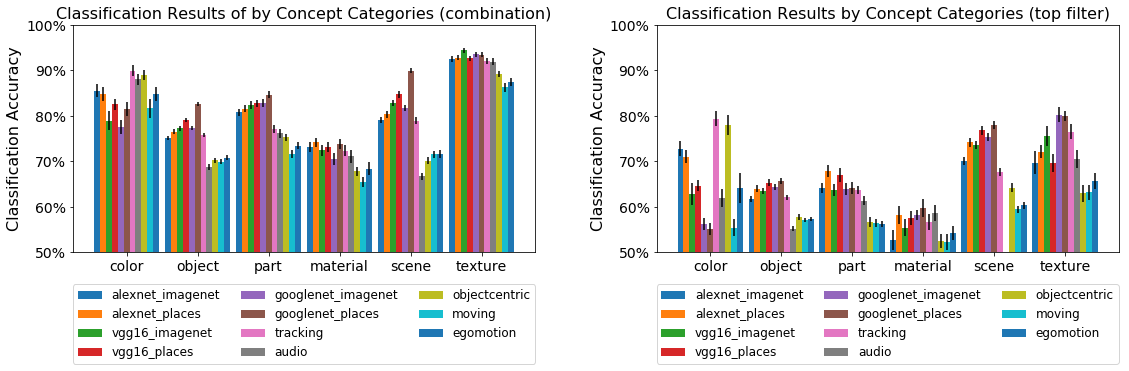}
  \caption{Segmentation (top) and classification (bottom) results for additional networks \& datasets.}
  \label{f:additional_results}
\end{figure}

\Cref{f:additional_results} shows segmentation (top) and classification (bottom) results when using AlexNet (AN) conv5, VGG16 (VGG) conv5\_3, and GoogLeNet (GN) inception5b trained on both ImageNet (IN) and Places365 (P) as well as conv5 of these self-supervised (SS), AlexNet networks: tracking, audio, objectcentric, moving, and egomotion. GN performed worse than VGG because of its lower spatial resolution ($7 \times 7$ vs. $14 \times 14$); GN-IN inception4e ($14 \times 14$) outperforms VGG-IN conv5\_3 (supp. fig. 11). In~\cite{bau2017network}, GN detects scenes well, which we exclude due to lack of segmentation data. SS performance improves more than supervised networks (5-6x vs. 2-4x), suggesting that SS networks encode BRODEN concepts more distributedly. 
\section{Interpretability}\label{s:interp}

In this section, we propose a new standard for visualizing non-extreme examples, show how the single- and multi-filter perspectives can be unified, and demonstrate how viewing concept weights as embeddings in filter space give us novel explanatory power.

\subsection{Visualizing Non-Maximal Examples}\label{s:nonmax}
Many visual explanation methods demonstrate their value by showing visualizations of inputs that maximally activate a filter, whether that be real, maximally-activating image patches~\cite{Zeiler13}; learned, generated maximally-activated inputs~\cite{Mahendran15,nguyen2016synthesizing}; or filter segmentation masks for maximally-activating images from a probe dataset~\cite{bau2017network}. 

\begin{figure}
\begin{center}
\includegraphics[width=0.50\linewidth]{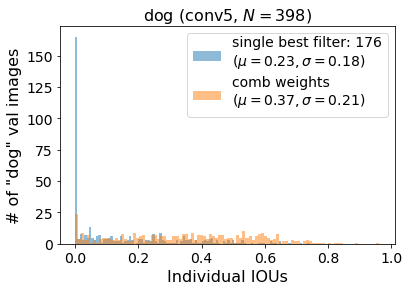}
\includegraphics[width=0.48\linewidth]{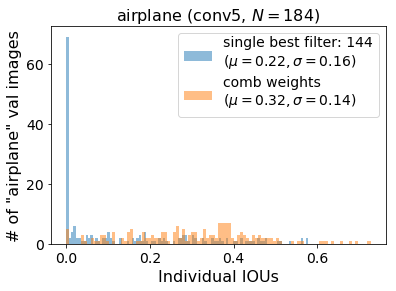}
\end{center}
\caption{The empirical $\IoUind$ distribution when using the best single filter and the learned weights for `dog' (left) and `train' (right) ($\mu, \sigma$ computed on the non-zero part of each distribution).}
\label{f:dist_ind_ious}
\end{figure}

While useful, these approaches fail to consider how visualizations differ across the distribution of examples.~\Cref{f:dist_ind_ious} shows that using a single filter to segment concepts~\cite{bau2017network} yields $\IoUind$ scores of $0$ for many examples; such examples are simply not considered by the set IoU metric. This often occurs because no activations survive the $\tau$-thresholding step, which suggests that a single filter does not consistently fire strongly on a given concept.

\begin{figure*}
\centering
\includegraphics[width=\linewidth]{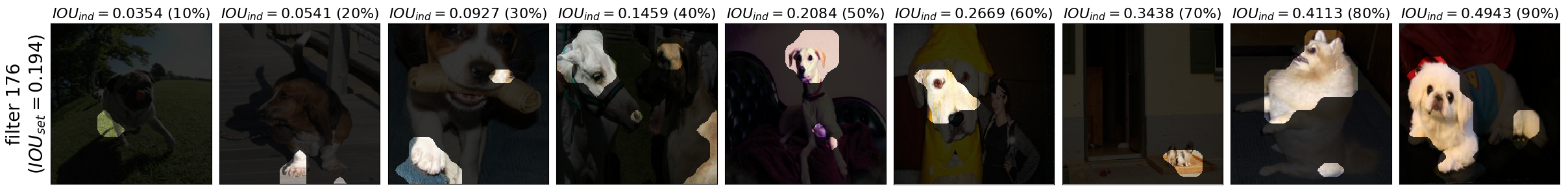}
\includegraphics[width=\linewidth]{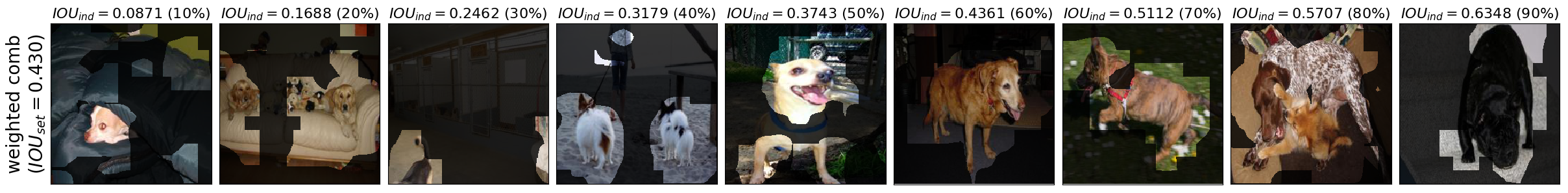}
\includegraphics[width=\linewidth]{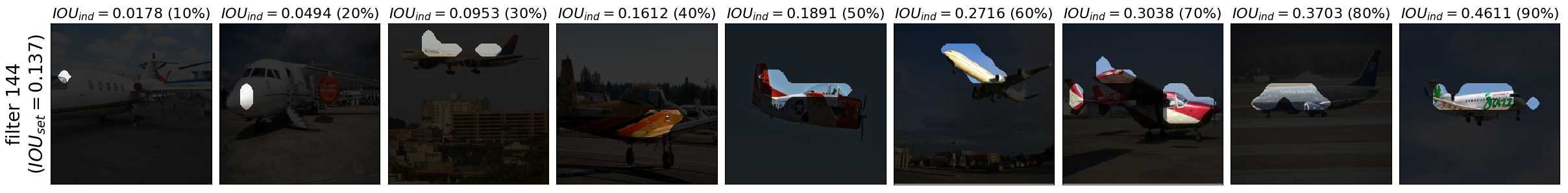}
\includegraphics[width=\linewidth]{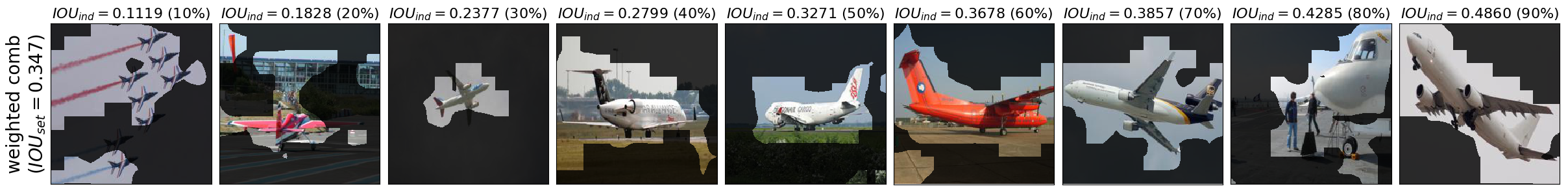}
\caption{For the `dog' and `airplane' concepts, an example is automatically selected at each decile of the non-zero portion of the distribution of individual IoU scores (\Cref{f:dist_ind_ious}), and the predicted conv5 segmentation masks using the best filter (odd rows) as well as the learned weights (even rows) are overlaid.}
\label{f:viz_dist}
\end{figure*}

We argue that a visualization technique should still work on and be informative for non-maximal examples. In~\Cref{f:viz_dist}, we automatically select and visualize examples at each decile of the non-zero portion of the individual IoU distribution (\cref{f:dist_ind_ious}) using both learned concept weights and the best filters identified for each of the visualized categories. For `dog' and `airplane' visualizations using our weighted combination method, the predicted masks are informative and salient for most of the examples, even the lowest 10th percentile (leftmost column). Ideally, using this decile sampling method, the visualizations should appear salient even for examples from lower deciles. However, for examples using the best single filter (odd rows), the visualizations are not interpretable until higher deciles (rightmost columns). This is in contrast to the visually appealing, maximally activating examples shown in supp. fig. 13.

\begin{table*}
\centering
\caption{Nearest concepts (in $\cos$ distance) using segmentation (left sub-columns) and classification (right) conv5 embeddings.}
\label{t:nearest_neighbors}
\begin{tabular}{|c|c|c|c|c|c|c|c|c|c|}
\hline
\multicolumn{2}{|c|}{\fns{\textbf{dog}}} & \multicolumn{2}{c|}{\fns{\textbf{house}}} & \multicolumn{2}{c|}{\fns{\textbf{wheel}}} & \multicolumn{2}{c|}{\fns{\textbf{street}}} & \multicolumn{2}{c|}{\fns{\textbf{bedroom}}} \\ \hline \hline
\fns{cat (0.81)}&\fns{muzzle (0.73)}&\fns{building (0.77)}&\fns{path (0.56)}&\fns{bicycle (0.86)}&\fns{headlight (0.66)}&\fns{n/a}&\fns{sidewalk (0.74)}&\fns{n/a}&\fns{headboard (0.90)}\\
\fns{horse (0.73)}&\fns{paw (0.65)}&\fns{henhouse (0.62)}&\fns{dacha (0.54)}&\fns{motorbike (0.66)}&\fns{car (0.53)}&\fns{n/a}&\fns{streetlight (0.73)}&\fns{n/a}&\fns{bed (0.85)}\\
\fns{muzzle (0.73)}&\fns{tail (0.52)}&\fns{balcony (0.56)}&\fns{hovel (0.54)}&\fns{carriage (0.54)}&\fns{bicycle (0.52)}&\fns{n/a}&\fns{license plate (0.73)}&\fns{n/a}&\fns{pillow (0.84)}\\
\fns{ear (0.72)}&\fns{nose (0.47)}&\fns{bandstand (0.54)}&\fns{chimney (0.53)}&\fns{wheelchair (0.53)}&\fns{road (0.51)}&\fns{n/a}&\fns{traffic light (0.73)}&\fns{n/a}&\fns{footboard (0.82)}\\
\fns{tail (0.72)}&\fns{torso (0.44)}&\fns{watchtower (0.52)}&\fns{earth (0.52)}&\fns{water wheel (0.48)}&\fns{license plate (0.49)}&\fns{n/a}&\fns{windshield (0.71)}&\fns{n/a}&\fns{shade (0.74)}\\ \hline
\end{tabular}
\end{table*}

\setlength{\tabcolsep}{2pt}
\begin{table}
\centering
\caption{Vector arithmetic using segmentation, conv5 weights.}
\label{t:vec_math}
\begin{tabular}{|c|c|c|c|}
\hline
\fns{\textbf{grass $+$ blue $-$ green}} & \fns{\textbf{grass $-$ green}} & \fns{\textbf{tree $-$ wood}} & \fns{\textbf{person $-$ torso}} \\ \hline \hline
\fns{sky (0.17)}&\fns{earth (0.22)}&\fns{plant (0.36)}&\fns{foot (0.12)}\\\hline
\fns{patio (0.10)}&\fns{path (0.21)}&\fns{flower (0.29)}&\fns{hand (0.10)}\\ \hline
\fns{greenhouse (0.10)}&\fns{brown (0.18)}&\fns{brush (0.29)}&\fns{grass (0.09)}\\ \hline
\fns{purple (0.09)}&\fns{sand (0.16)}&\fns{bush (0.28)}&\fns{mountn. pass (0.09)}\\ \hline
\fns{water (0.09)}&\fns{patio (0.15)}&\fns{green (0.25)}&\fns{backpack (0.09)}\\ \hline
\end{tabular}
\end{table}


\subsection{Unifying Single- \& Multi-Filter Views}
\begin{figure}
\centering
\includegraphics[width=0.49\linewidth]{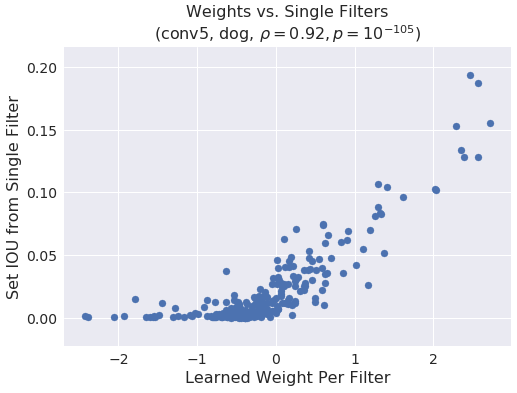}
\includegraphics[width=0.49\linewidth]{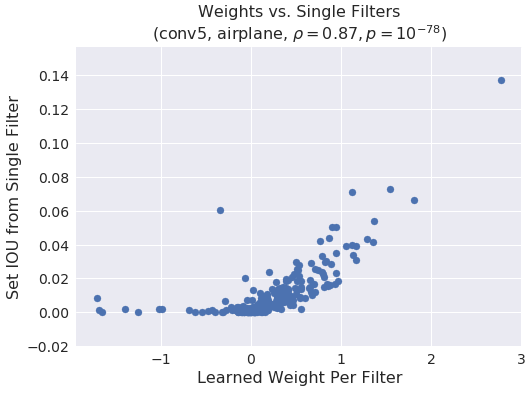}
\caption{Correlation between learned segmentation weights and each filter's set IoU score for `dog' (left) and `airplane' (right).}
\label{f:correlation_weights_single}
\end{figure}

\Cref{f:correlation_weights_single} highlights that single filter performance is often strongly, linearly correlated with the learned weights $\bw$, thereby showing that individual filter performance is indicative of how weighted it'd be in a linear filter combination. Visually, a filter's set IoU score appears correlated with its associated weight value passed through a ReLU, i.e., $\max(w_k,0)$. For each of the $682$ BRODEN segmentation concepts and each AlexNet layer, we computed the correlation between $\max(\bw,0)$ and $\{\IoUset(c;M_k,\text{val})\}_{k=1\ldots K}$. By conv3, around $80\%$ of segmentation concepts are significantly correlated ($p < 0.01$): conv1: 47.33\%, conv2: 69.12\%, conv3: 81.14\%, conv4: 79.13\%, conv5: 82.47\%. Thus, we show how the single filter perspective can be unified with and utilized to explain the distributive perspective: we can quantify how much a single filter $k$ contributes to concept $c$'s encoding from either $\frac{|w_{k}|}{\lVert \mathbf{w}\lVert_1}$ where $\mathbf{w}$ is $c$'s learned weight vector or $\frac{\text{IoUset}(c; M_{k^*},\text{val})}{\text{IoUset}(c; M(\cdot;\mathbf{w}),\text{val})}$.


\subsection{Explanatory Power via Concept Embeddings}
Finally, the learned weights can be considered as embeddings, where each dimension corresponds to a filter. Then, we can leverage the rich literature~\cite{mikolov2013distributed,mikolov2013linguistic,kiros2014unifying} on word embeddings derived from textual data to better understand which concepts are similar to each other in network space. To our knowledge, this is the first work that learns semantic embeddings aligned to the filter space of a network from visual data alone. (For this section, concept weights are normalized to be unit length, i.e., $\bw' = \frac{\bw}{\lVert\bw\rVert}$).

~\Cref{t:nearest_neighbors} shows the five closest concepts in cosine distance, where $1$ denotes that $\bw_1'$ is $\ang{0}$ from $\bw_2'$ and $-1$ denotes that $\bw_1'$ is $\ang{180}$ from $\bw_2'$. These examples suggest that the embeddings from the segmentation and classification tasks capture slightly different relationships between concepts. Specifically, the nearby concepts in segmentation space appear to be similar-category objects (i.e., animals in the case of `cat' and `horse' being nearest to `dog'), whereas the nearby concepts in classification space appear to be concepts that are related compositionally (i.e., parts of an object in the case of `muzzle' and `paw' being nearest to `dog'). Note that `street' and `bedroom' are categorized as scenes and thus lack segmentation annotations.

\paragraph{Understanding the Embedding Space.}\Cref{t:vec_math} shows that we can also do vector arithmetic by adding and subtracting concept embeddings to get meaningful results. For instance, we observe an analogy relationship between `grass'$-$`green' and `sky'$-$`blue' and other coherent results, such as non-green, `ground'-like concepts for `grass' minus `green' and floral concepts for `tree' minus `wood'. t-SNE visualizations and K-means clustering (see supp. table 2 and supp. figs. 16 and 17) also demonstrate that networks learn meaningful, semantic relationships between concepts.

\begin{figure}[th]
  \centering
  \includegraphics[width=0.59\linewidth]{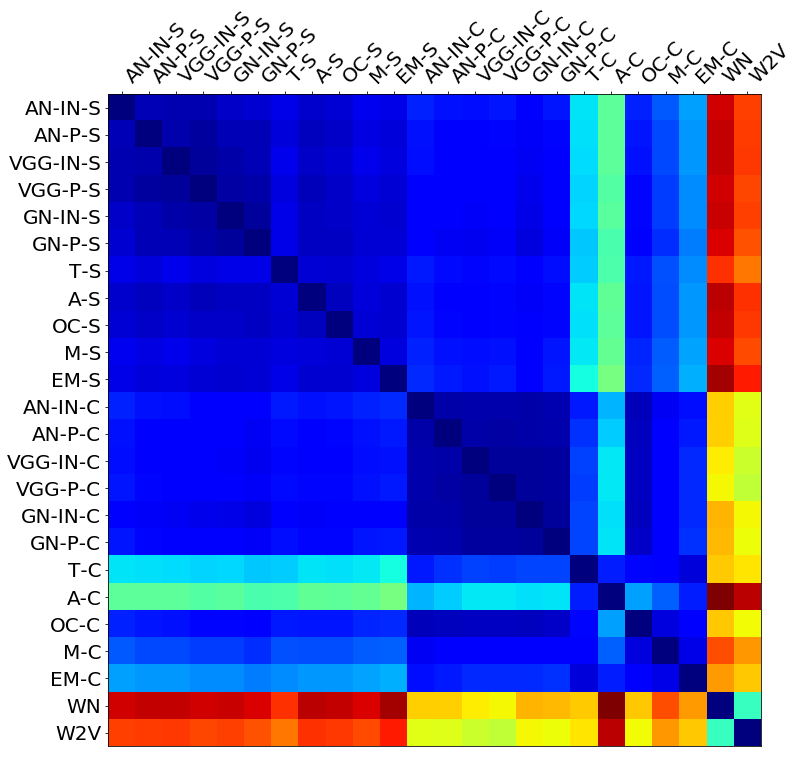}
  \includegraphics[width=0.39\linewidth]{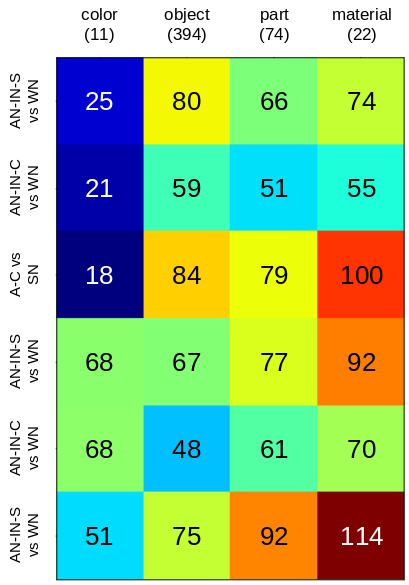}
  \caption{Comparing Net2Vec embeddings quantitatively. Left: Each cell corresponds to distance $D_{i,j}$ for embedding spaces $i$ and $j$ (see~\cref{s:more_experiments} for abbreviations). Right: Each cell corresponds to mean distance $D_{i,j,c}$ for each concept category.}
  \label{f:net2vec_analysis}
\end{figure}

\paragraph{Comparing Embeddings from Different Learned Representations.} The learned embeddings extracted from individual networks can be compared with one another quantitatively (as well as to other semantic representations). Let $d(W): \mathbb{R}^{C \times K} \to \mathbb{R}^{C \times C}= W \cdot W^T$ compute the cosine distance matrix for $C$ concepts of a given representation (e.g., AlexNet), whose normalized embeddings $\bw'$ form the rows of $W$. Then, $D_{i,j} = \lVert d(W^i) - d(W^j) \rVert_2^2$ quantifies the distance between two embedding spaces $W^i, W^j$, and $D_{i,j,c} = \lVert d(W^i)_c - d(W^j)_c \rVert_2^2$ does that for concept $c$.~\Cref{f:net2vec_analysis} (left) shows $D_{i,j}$ between 24 embedding spaces: 2 tasks $\times$ 11 network, WordNet (WN), and Word2Vec (W2V) ($C = 501$, the number of BRODEN concepts available for all embeddings; see supp. sec. 3.2.1). It shows that tracking and audio (T, A) classification embeddings are quite different from others, and that classification embeddings (-C) are more aligned to WN and W2V than segmentation ones (-S).~\Cref{f:net2vec_analysis} (right) shows select mean $D_{i,j,c}$ distances averaged over concept categories. It demonstrates that colors are quite similar between WN and network embeddings and that materials most differ between audio and the WN and W2V embeddings.

\section{Conclusion}
We present a paradigm for learning concept embeddings that are aligned to a CNN layer's filter space. Not only do we answer the binary questions, ``does a single filter encode a concept fully and exclusively?,'' we also introduce the idea of filter and concept ``overlap'' and outline methods for answering the scalar extension questions, ``to what extent...?'' We also propose a more fair standard for visualizing non-extreme examples and show how to explain distributed concept encodings via embeddings. While powerful and interpretable, our approach is limited by its linear nature; future work should explore non-linear ways concepts can be better aligned to the filter space.

\blfootnote{\noindent\textbf{Acknowledgements.} We gratefully acknowledge the support of the Rhodes Trust for Ruth Fong and ERC 677195-IDIU for Andrea Vedaldi.}
{\small\bibliographystyle{ieee}\bibliography{egbib,shortstrings,vgg_local,vgg_other}}
\end{document}